\documentclass[11pt,a4paper]{article}
\usepackage[T1]{fontenc}
\usepackage[utf8]{inputenc}
\usepackage{lmodern}
\usepackage[margin=15mm,footskip=9mm]{geometry}
\usepackage{amsmath,amssymb,graphicx,booktabs,array}
\usepackage{microtype}
\usepackage[numbers,super,sort&compress]{natbib}
\usepackage{xurl}
\usepackage[hidelinks]{hyperref}
\usepackage{bookmark}
\usepackage[font=small,labelfont=bf]{caption}
\hypersetup{
  pdftitle={Identifiable measurements of geometric action reasoning},
  pdfauthor={Jingyao Zhang, Yun Li, Lu Han},
  pdfsubject={Main manuscript, Experiment Process Atlas, and Supplementary Information},
  bookmarksopen=false,
  pdfdisplaydoctitle=true
}
\begin{document}
\pdfbookmark[0]{Main manuscript}{part.main}
\title{GAGR-Lab: Evaluating Joint Spatial-Geometric and Analytic Function Reasoning}
\author{Jingyao Zhang \and Yun Li \and Lu Han}
\date{}
\maketitle
\begin{abstract}
Executable geometric tasks connect symbolic answers to verifiable actions, while their interpretation depends on information access and task construction. We retrospectively audit 2,333 saved model-call records and 30 historical game episodes. Matched requests yield 30 formalization-success/synthesis-failure pairs and two reverse pairs. Among 176 valid/invalid candidate pairs, 89 reject both candidates. Exact checks show that every negative admits an endpoint counterexample. In 67 determinate later-stage support pairs, correct algebra and irrelevant support each succeed 21 times. All 187 image-arm requests also include complete symbolic geometry. A separate exact control finds one curve satisfying 72 distinct catalog scenes. Archived screenshots, complete selected episodes and dense trajectories make these observations inspectable. An elementary quadratic certificate supplies a conditional explanation for an earlier reported failure family. These results motivate identifiable measurements and prospective tests of transfer between formalization, construction and verification.
\end{abstract}
A spatial-to-function task asks a model to construct a symbolic object whose executed geometry satisfies a public contract. The resulting curve can be checked against endpoints, obstacles, regularity and global bounds. This supplies direct behavioral evidence about the connection between geometric descriptions and mathematical actions. A scientific evaluation additionally needs to identify which information and obligations are necessary for success.

GAGR-Lab provides a Graphwar-derived native arena and public mathematical scene contracts. The setting connects spatial relations to executable functions. Later framework versions retained multi-configuration requests, raw model outputs and fixed scoring records. We analyze these saved observations together with three historical native-game cohorts. Earlier pilot claims and their outstanding source requirements are retained in the Supplementary Information.

We focus on three measurement questions. Which mathematical obligations explain a failed action? Which behavioral differences are visible when the same scene and configuration are matched across operations? Which successful behaviors actually require the information supplied by a visual stimulus? These questions motivate the paired judgments, support comparisons, trace reconstructions and constructive controls reported below.

The present contribution is a retrospective evidence study. Its core observations come from source-hashed model outputs. Its mathematical controls provide externally checkable statements about task construction. Prospective experiments in the discussion specify how the associated mechanistic hypotheses can be tested.

MathVista studies mathematical reasoning in visual contexts \cite{ref1}, and DynaMath uses generated variants to examine robustness \cite{ref2}. SpaceCast-Bench organizes predictive spatial questions around scene observation, transformation and inference \cite{ref3}. Our action space consists of executable functions or mathematical constraint objects, enabling explicit counterexamples for endpoint and continuous-geometry obligations. The arena setting also preserves the images and outcomes available at each turn.

Metric choice can affect apparent capability patterns \cite{ref4}. This concern extends to the information supplied to a model, the construction of candidate answers and the structure of admissible solution sets. Exact certificates and inspectable source data allow these conditions to be evaluated explicitly.

Generation and verification have already been compared across reasoning domains \cite{ref5,ref6}. Our contribution concerns executable continuous geometry, exact candidate counterexamples and source-bound action traces. The question is how operation-level differences survive checks on input access, candidate construction and proof coverage.

\section*{Results}
\subsection*{Source-bound archive coverage}
The three native cohorts contain 6, 12 and 12 completed episodes. Their raw turn events and game summaries agree on 34, 180 and 221 action attempts, with 26, 141 and 9 engine-valid trajectories. They record four, four and one hits. The second cohort was recovered from its original games.zip. Selected game manifests were checked against the actual screenshot, log and trajectory bytes. These cohorts use different requested model identifiers, prompts, temperatures and output limits.

The later saved-call archive contains 1,341 main observations, 968 historical observations and 24 complex formalization observations, with 2,333 unique archived attempt identities. We rehash every original response and associated score record. The main set comprises 45 qualification requests, 15 syntax-qualification requests, 751 exploration requests and 530 requests from a later frozen plan named confirmation in the archive. That plan completed 530 of 2,016 scheduled jobs. Exploration completed 751 of 1,988 scheduled jobs. Unexecuted positions remain part of the coverage record.

Exploration and the later stage include 14 requested configurations across nine base-model labels. Qualification also contains an additional default-mode label. Requested ON, OFF and DEFAULT modes retain their archived meanings. Serving identities and backend snapshots are limited to the recorded provider metadata. Shared scenes and repeated configurations are dependent observations. We present counts and within-block comparisons without treating requests as independent scene samples.

A matched block has the same archived stage, profile identifier and world identifier. The main observations cover 23 distinct world identities, including 10 FOUNDATION, 9 TRANSITION and 4 ADVANCED design labels. These labels describe the generator configuration. The numbers of fully shared candidate-judgment scenes across all 14 configurations are five in exploration and one in the later stage. This common panel is retained explicitly.

Scores in this analysis use the saved VGAGR4131\_20261105\_1 regrading epoch. TRUE and FALSE denote determinate mathematical judgments in the supported fragment. NOT\_EVALUATED records interface or other pre-mathematical failures. UNKNOWN preserves proof-coverage or resource limits. The main set contains 673 TRUE, 570 FALSE, 77 NOT\_EVALUATED and 21 UNKNOWN results. The analysis added zero model API calls and zero model observations.

Supplementary Tables 1-3 report the scoring-state counts, requested serving profiles and historical native-cohort denominators. They retain unresolved observations and incomplete plan coverage.

\subsection*{Formalization and construction separate in matched requests}
Among same-stage, same-profile, same-world formalization and direct synthesis requests, 30 pairs have correct formalization and failed synthesis, split into 16 exploration pairs and 14 later-stage pairs. Two exploration pairs show the reverse combination. Fifty-one pairs succeed on both operations, and 27 fail on both (Fig. 1a,b). The 30 formalization-success/synthesis-failure pairs span 12 world identities, 11 configurations and seven requested model labels. These observations describe separate requests under their operation-specific output contracts.

Among 406 failed polynomial synthesis outputs with exact coefficient diagnostics, 142 satisfy both endpoint equalities (Fig. 1c). Their failure therefore concerns other contract obligations. The remaining 264 violate at least one endpoint equality and may also violate other requirements. These diagnostics motivate a staged analysis of endpoint satisfaction and whole-domain feasibility.

\begin{figure}[htbp]\centering
\includegraphics[width=180mm]{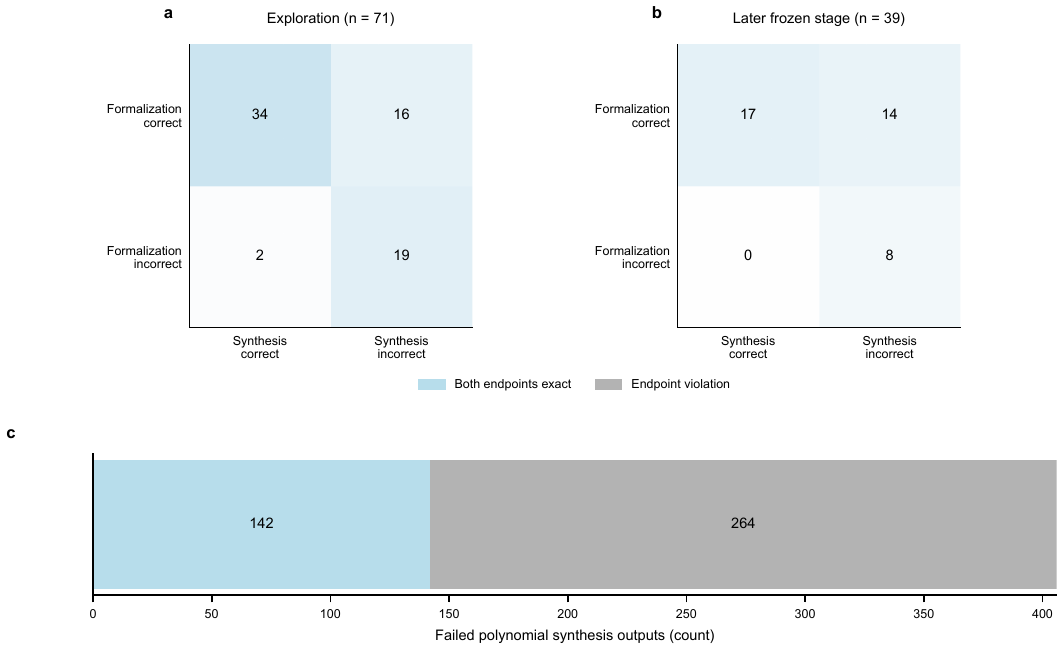}
\caption{Matched formalization, construction and endpoint outcomes. a,b, Formalization and direct synthesis correctness in exploration (a) and the later frozen stage (b). The matrices contain 71 and 39 jointly determinate stage-profile-world pairs. An additional 35 and 33 pairs contain an unevaluated or unknown result. c, Endpoint classification within 406 failed polynomial synthesis outputs covered by the exact diagnostic. Blue-tinted bars denote two exact endpoints, and grey denotes at least one endpoint violation. Values are descriptive counts of dependent archived observations.}
\end{figure}
\subsection*{Candidate construction conditions judgment patterns}
Judgment responses reveal a structured asymmetry. Of 176 positive/negative pairs, 173 have determinate scores for both requests. Sixty-six pairs classify both candidates correctly, 89 reject both, 16 accept both and two classify both incorrectly. The remaining three pairs have unresolved outputs. In the later frozen stage, the corresponding counts are 27, 39, five and zero among 71 pairs. The shared panel and all determinate pairs are shown in Fig. 2a,b.

The answer pattern depends on the scene family. BASIC\_ENDPOINTS has seven both-correct, nine accept-both and two both-wrong outcomes across 18 pairs. FREE\_BYPASS has 12 both-correct and 19 reject-both outcomes across 31 pairs. These are associations within the recorded design. Scene structure, expression form and instruction length co-vary in the archived families.

Exact endpoint evaluation exposes an important property of the judgment probes. Every negative candidate in the 18 paired worlds violates at least one endpoint equality. Consequently, all 176 negative requests admit an endpoint-only counterexample. Certifying the corresponding positive requires the complete public contract. Rejection accuracy on these negatives is compatible with a local falsification procedure. A targeted global-geometry measurement requires negative candidates that preserve endpoints and all non-target obligations.

A concrete FREE\_BYPASS pair differs only by 1/1,000 in the linear coefficient (Fig. 2c,d). The archived qwen3.7-flash OFF configuration rejects both candidates. Both candidates are globally smooth polynomials. The negative violates the endpoint equalities. This case makes the candidate construction and the model's two binary answers inspectable together.

\begin{figure}[htbp]\centering
\includegraphics[width=180mm]{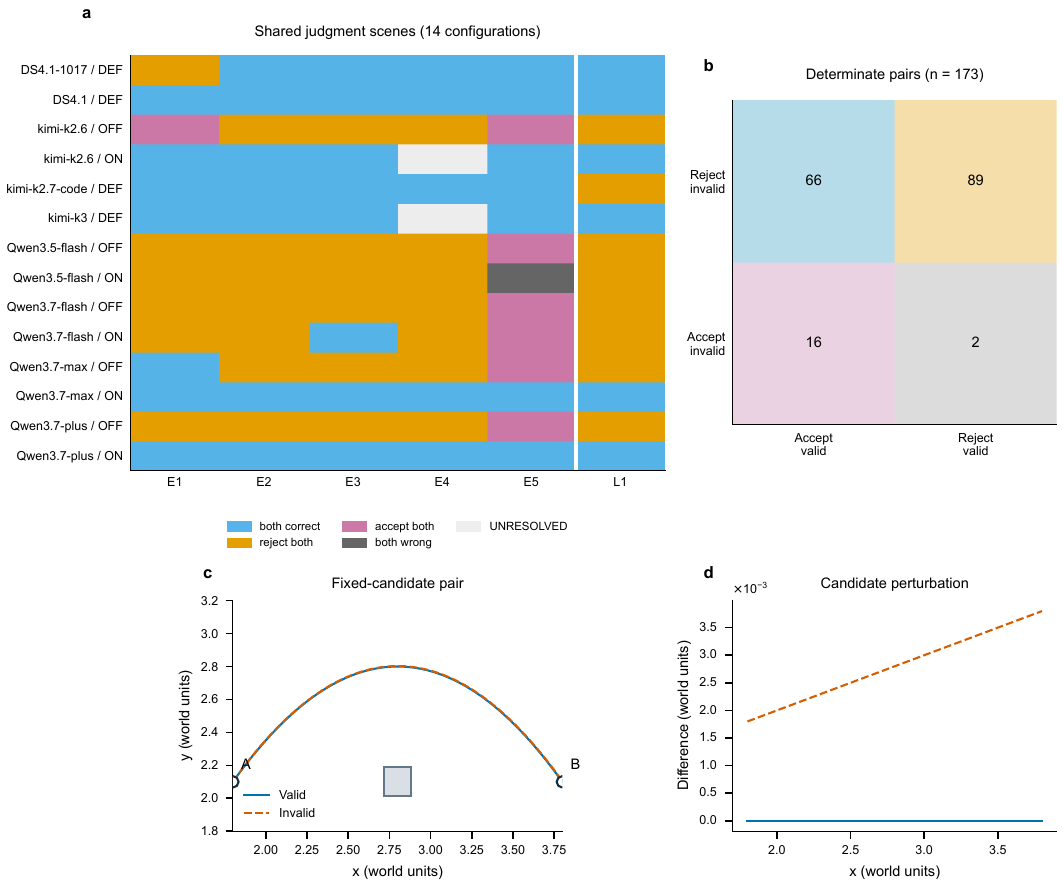}
\caption{Paired judgments and a source-bound counterexample. a, The 84 configuration-scene pairs shared by all 14 configurations, covering five exploration scenes (E1-E5) and one later-stage scene (L1). Colours encode the joint answers to valid and invalid candidates. b, All 173 jointly determinate candidate pairs, with three further unresolved pairs outside the matrix. c, A local geometry view of the certified valid polynomial and the endpoint-violating candidate in one FREE\_BYPASS scene. Solid blue and dashed orange identify the candidates. d, Their difference from the valid polynomial. The difference panel has a magnified vertical scale. Full scene contracts, candidate expressions and model answers accompany the source data. DEF denotes the DEFAULT requested mode.}
\end{figure}
\subsection*{Correct symbolic support has limited net advantage}
Correct symbolic support shows little net advantage in the retained matched data. There are 178 correct-support versus irrelevant-support pairs with identical non-message request parameters. The correct-support text is consistently 54 characters shorter. In exploration, 102 pairs have determinate scores in both arms. Correct support succeeds alone in four and irrelevant support succeeds alone in one. In the later stage, 67 pairs are jointly determinate, with 21 successes in each arm and three discordant wins for each arm. Nine pairs across the stages contain unresolved outcomes. Among 167 pairs with exact polynomial outputs in both arms, 49 return the same polynomial.

The paired outcomes are shown separately for exploration and the later stage (Fig. 3a,b). Both conditions already contain complete geometric descriptions. Equal observed success counts in the later-stage subset leave the utility of alternative representations, non-redundant support and token-controlled interventions open.

\begin{figure}[htbp]\centering
\includegraphics[width=180mm]{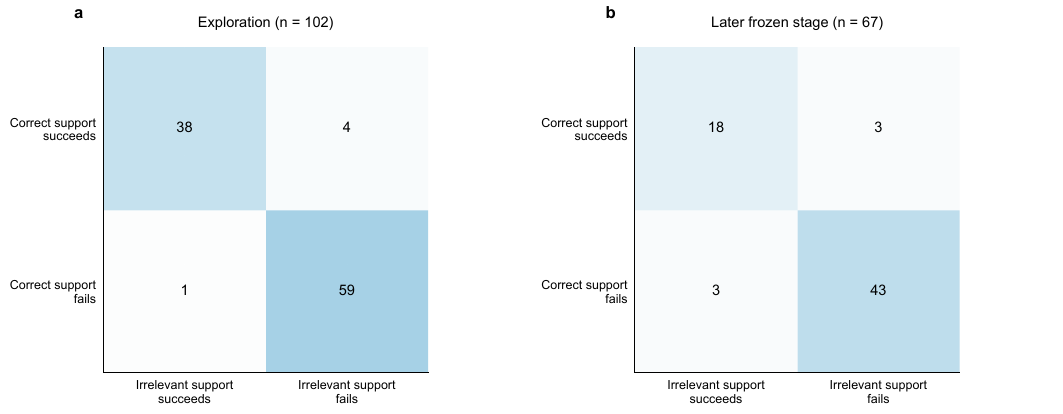}
\caption{Matched correct-support and irrelevant-support outcomes. a, Exploration contains 102 jointly determinate stage-profile-world pairs and four additional unresolved pairs. b, The later frozen stage contains 67 jointly determinate pairs and five additional unresolved pairs. Rows indicate correct-support success or failure, and columns indicate irrelevant-support success or failure. Numbers are paired counts. Both arms include complete geometric information and identical non-message request parameters. Correct-support text is 54 characters shorter in every pair.}
\end{figure}
\subsection*{Information access and answer-set necessity}
Actual request identities support a direct audit of observation conditions. All 1,341 main response records match their frozen request-body hashes. Each of the 187 requests labeled SYN\_IMAGE contains one image attachment and complete symbolic scene geometry, including endpoints, canvas, objects, constraints and function space. The appropriate condition label is image plus complete symbolic geometry. This supplies a clear basis for designing a future image-only condition.

A separate finite catalog demonstrates why instance information must be necessary for the scored operation. Its 72 scenes vary circle or rectangle geometry, obstacle position, gap width and wall shift. A single fixed piecewise-linear curve satisfies every scene. We reconstruct the frozen scene identities, verify the original candidate hash and rerun the fixed exact geometry oracle on all 72 scenes. All checks pass. The catalog shares one base scene family.

The input audit and two distinct catalog instances are shown in Fig. 4a-c. The common curve has vertices (0,0), (1/8,1), (3/4,1) and (1,0).

Let A(w) be the admissible answer set for instance w. A common answer in the intersection over instances supplies an instance-independent success policy. A benchmark intended to require instance-specific visual information should therefore audit this intersection, alongside the mapping from pixels to geometry. In the same catalog, one fixed complete geometry-readout answer matches only one of the 72 scenes. Synthesis and full readout thus present different necessity conditions.

\begin{equation}\bigcap_{w\in\mathcal{W}}\mathcal{A}(w)\ne\varnothing\end{equation}
\begin{figure}[htbp]\centering
\includegraphics[width=180mm]{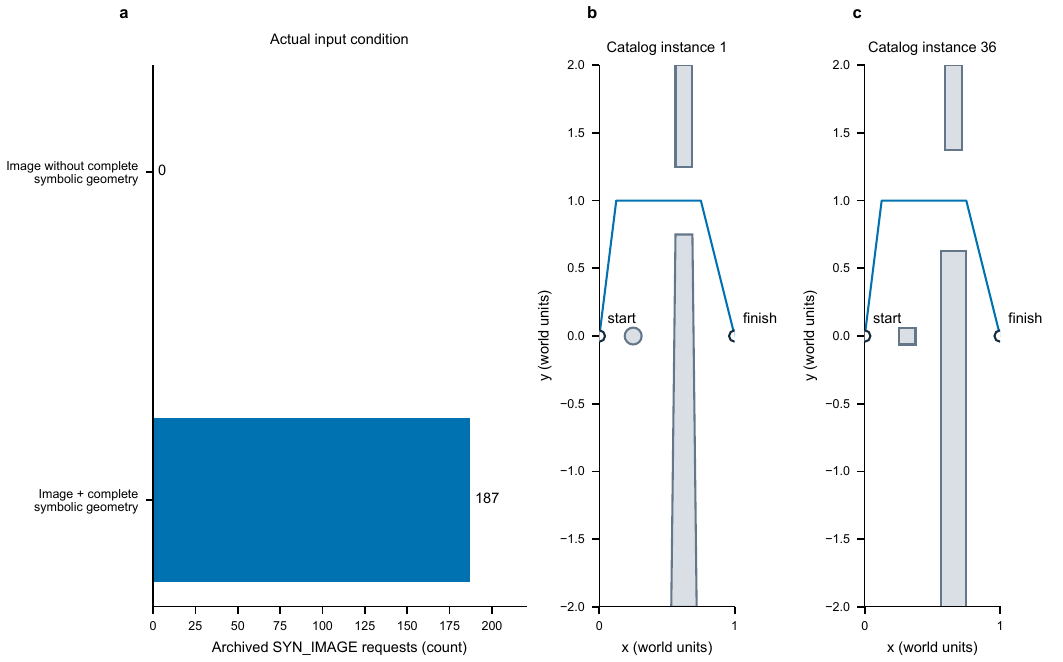}
\caption{Actual information conditions and a common catalog solution. a, Input-content classification of all 187 archived SYN\_IMAGE requests. Each contains an image and complete symbolic geometry. b,c, Instances 1 and 36 of the 72-instance frozen visual catalog, drawn at equal horizontal and vertical world-unit scales. The solid blue curve is the same in both panels and passes fixed exact geometric checks in all 72 instances. Grey shapes denote obstacles. The catalog is one base scene family and this control contains no new model observations.}
\end{figure}
\subsection*{Archived native actions expose concrete failures and successes}
Dense archived native paths provide concrete interaction-level evidence. We select the earliest completed hit episode from each historical cohort for inspection. The selected cases contain seven turns in total. The casebook preserves every selected turn, including the actual screenshot, raw visible response, historical executed expression, outcome fields and full CSV. The displayed hit prefix ends at the first stored sample inside the target region, with an option to display the complete stored continuation.

In the first case, a scene labeled hard contains nine obstacles. The first actor returns y = -25.92*sin(x), and the historical extraction policy executes -25.92*sin(x). Its path exits the world. The second actor executes 0.235*x and records a hit. A linear action therefore succeeds in this particular nine-obstacle scene. Obstacle count needs to be supplemented by active blocking relations and necessary solution complexity when defining a calibrated difficulty scale.

The second case contains four turns. An early quadratic exits the world, while the later expression 1-18*(x-17)\textasciicircum{}2/1681 records a hit. The third case records a first-turn hit with -0.633*x-5.31. These paths retain their historical engine and protocol provenance. The examples provide inspectable successful and unsuccessful actions under the recorded configurations.

Fig. 5a-c shows the first failed action and the recorded hit where both are available. Supplementary Figs. 1-3 retain every turn and its actual input screenshot. Supplementary Tables 4-6 preserve the exact visible answers and historical executed expressions.

\begin{figure}[htbp]\centering
\includegraphics[width=180mm]{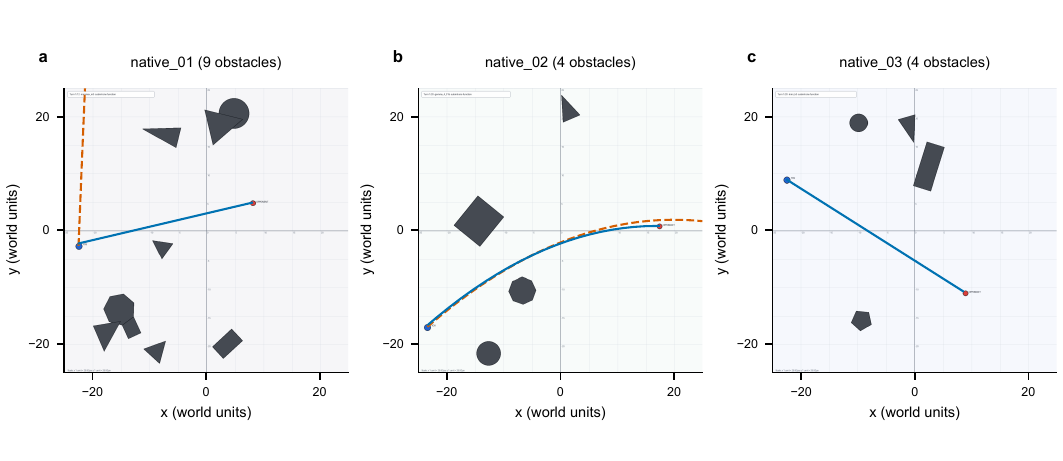}
\caption{Source-bound native action traces in three selected episodes. a-c, The earliest completed hit episode from each historical cohort. Each panel uses that episode's first-turn input image. Dashed orange traces show the first failed action where available, and solid blue traces show the recorded hit prefix. Panel a contains nine obstacles, and panels b and c contain four each. The panels are selected cases from 30 completed games with 435 actions. Their comparative scope is episode inspection. Full per-turn images and stored CSV continuations accompany the Supplementary Information.}
\end{figure}
\subsection*{An analytical constraint on quadratic-family feasibility}
The earlier draft reports that all 429 valid ordinary responses in its 72-game pilot begin with a unit quadratic term. We derive an elementary range certificate that would make this family unsuitable for its stated scene geometry. The corresponding 72-game raw log directory remains to be located, so the application to those reported responses retains a conditional provenance status.

Consider q(x)=a x squared+b x+c on an interval with horizontal span L. If its values stay in a vertical strip of height H throughout that interval, the following necessary bound holds. It is tight when slope and vertical offset are free.

\begin{equation}|a|L^2\leq4H\end{equation}
To see this, evaluate the polynomial at both endpoints and the midpoint. The absolute value of the resulting second difference is at most twice the strip height. Substitution gives the bound. A parabola with equal endpoint values and its vertex at the midpoint attains equality for its own range.

\begin{equation}q(u)+q(v)-2q((u+v)/2)=a(v-u)^2/2\end{equation}
Under the earlier manuscript settings, a center separation of at least 26 and launch and target radii bounded by one half require at least 25 horizontal units of travel. With H=50, a unit quadratic has minimum vertical range 625/4=156.25, exceeding the world height. The necessary absolute coefficient ceiling is 8/25. Automatic vertical anchoring changes the offset and preserves this range argument. The proof applies to continuous quadratic graphs over the required interval.

This certificate suggests a direct follow-up on the earlier pilot. Once its raw archive is located, every accepted response can be checked for its actual coefficient, traversed span and range bound. Additive constants can also be quotiented under the native anchoring law, allowing expression diversity to be separated from path diversity. C3 regularity and polynomial degree retain distinct definitions throughout these checks.

The coefficient bound and a constructed example are shown in Fig. 6a,b. The raw earlier pilot is still required to determine which complete returned functions satisfy the quadratic-family premise.

\begin{figure}[htbp]\centering
\includegraphics[width=180mm]{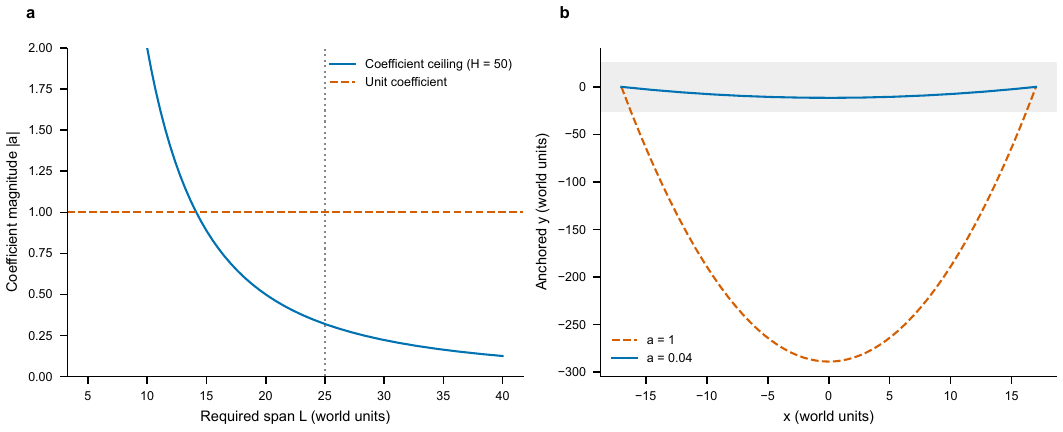}
\caption{Quadratic coefficient-span control. a, The necessary coefficient ceiling 4H/L squared for a vertical strip of height H = 50. The dashed orange line denotes a unit coefficient, and the dotted line marks a required span of 25 world units. b, Constructed continuous quadratic curves with coefficients 1 and 0.04. Grey shading denotes a strip of height 50. These are analytical controls. Their application to the earlier reported pilot retains the raw-source conditions described in the text.}
\end{figure}
\section*{Discussion}
The clearest model-level observation is an operation-specific separation between formalization and construction in matched requests. Correctly expressing relations can coexist with an unsuccessful constructed object. Correct symbolic support also has limited net benefit in the retained paired sample. Together, these results motivate a testable account of unstable transfer from explicit relations to executable, jointly constrained actions.

The archived call study is retrospective and incompletely executed. Its later frozen plan provides a temporally distinct subset, while the new analysis was developed after the outputs were available. The shared complete panel is small. Requests vary across modes, temperatures, output limits and serving conditions, and the same geometry is reused across configurations. The observed associations have the scope of these records.

The positive/negative construction needs a stronger intervention. A useful design would preserve endpoints and all non-target requirements while violating exactly one geometric or analytic property, then balance comparable perturbations across valid and invalid candidates. A separate equivalence intervention could rewrite the same function without changing its geometry. Such controls would test whether the observed answer patterns depend on syntax, numerical tolerance, proof burden or the target property.

The support study motivates a fully shared, length-controlled replication with input token counts and output budgets frozen. For vision, stimulus changes should alter admissible answers, and image-only input should have a qualified readout contract. Prospective scene families should receive solvability, necessity and single-factor difficulty certificates before model calls. Calibration and held-out validation should be grouped by base scene family.

These records identify behavioral outcomes. Internal representations and causal reasoning stages remain hypotheses. Differences in output contracts, local versus global proof obligations and serving conditions offer alternative explanations that can be separated by intervention. For example, replacing a generated constraint object with a certified one would test whether the construction deficit persists after the intermediate representation is fixed.

Archived geometric-action experiments already support instance-level scientific analysis. Paired judgments reveal answer-pattern differences whose interpretation depends on negative-candidate construction. Matched operations exhibit formalization-construction dissociations. Support controls and actual request inspection delimit the information available to each agent. A shared successful curve across a visual catalog establishes a constructive necessity test, while a quadratic range certificate supplies a conditional explanation for an earlier failure pattern. These results organize a concrete program of prospective interventions and more identifiable measurement.

Supplementary Table 7 connects the preliminary findings to controlled tests. The most direct next step is to pass the same intermediate formal object through a chained workflow and intervene separately on its correctness, candidate construction and verification. Supplementary Table 8 retains earlier reported experiments whose raw sources remain outstanding.

\section*{Methods}
\subsection*{Action semantics, contracts and certificates}
Native arena actions follow the archived anchoring convention. A function is translated vertically through the shooter, while its horizontal argument remains the global coordinate. Historical extraction policies are preserved alongside raw visible output. The later formal tasks use their declared endpoint and full-domain contracts directly. Their function values specify the public graph, so the two regimes retain distinct semantics.

\begin{equation}Y_f(x)=y_s+f(x)-f(x_s)\end{equation}
For the later tasks, G2A requests require an equivalent formal constraint description. SYN requests require a function. DEC\_POS and DEC\_NEG ask whether a fixed candidate satisfies all requirements. Correct-algebra support and irrelevant support are supplied in separate synthesis arms. Candidate correctness is evaluated mechanically. The stored primary certificates use rational Bernstein and Sturm checks in their admitted fragments, sharing the validated intermediate representation. Their independence therefore applies to arithmetic and sign certification within that representation.

We independently evaluate endpoint obligations with restricted exact polynomial arithmetic. The diagnostic accepts explicit rational polynomial expressions, evaluates the named endpoints and records exact residuals. Unsupported expressions retain their diagnostic coverage status. Official model scores remain linked to their saved certificates. We also verify that each actual judgment candidate equals the corresponding positive or negative candidate in its frozen plan.

In the formal task contracts, C3 denotes the class of functions with three continuous derivatives on the declared domain. Polynomial degree and regularity are recorded separately. The archived synthesis prompt advertises exact rational-polynomial certification up to degree 16. Mathematical FALSE, interface NOT\_EVALUATED and proof-coverage UNKNOWN retain their distinct meanings.

\subsection*{Pairing, configurations and statistical analysis}
Each pair is formed within the same archived stage, profile identifier and world identifier. Judgment pairs additionally select the predefined positive and negative candidates. Support pairs select correct-algebra and irrelevant-support synthesis arms. Formalization/synthesis pairs select G2A and SYN\_GEOMETRY. Every request remains a separate archived attempt. Determinate-pair matrices retain only pairs whose two operation scores are TRUE or FALSE, and report excluded states explicitly.

Across exploration and the later frozen stage, the 14 observed profiles request a 16,384-token output limit. Temperature is zero except for the kimi-k3 DEFAULT profile, which requests 1.0. ON, OFF and DEFAULT are requested serving-mode labels. They do not establish hidden implementation identity. TEXT\_ONLY is the response-carrier setting. It remains compatible with image attachments in user messages. Supplementary Table 2 lists the complete archived profile labels.

Descriptive contrasts preserve the matched blocks and complete-panel identities. We report no independent-request confidence interval or confirmatory significance claim. Repeated scenes and related profiles are dependent units. The later plan was frozen before its requests, while the present scientific hypotheses were developed retrospectively after both stages were available.

\subsection*{Native case selection and trajectory rendering}
The native authority is the archived Graphwar-derived Rust implementation \cite{ref7}. We reconcile completed episode summaries against turn\_completed events and verify the selected episode manifests. Case selection takes the earliest completed hit episode in each of three cohorts, ordered by archived episode identifier. Every turn in the selected episodes is retained. Rendering overlays original CSV coordinates on the exact screenshot supplied at that turn, using the archived world bounds. For a hit, the default display ends at the first stored sample inside the target hit region. The full CSV continuation remains available in the casebook. The plots reconstruct archived execution and introduce zero new engine or model trials.

\subsection*{Exact diagnostics and data availability}
The supplementary endpoint diagnostic parses a restricted arithmetic syntax tree with exact rational coefficients. It accepts explicit polynomial operations subject to node, degree and coefficient resource bounds. Unsupported diagnostic expressions retain their coverage status. No missing function, bracket, coefficient or model answer is inferred. DEC scores compare the returned binary judgment with the predefined certified candidate truth. All primary request-body hashes are checked against canonical serialization of their frozen jobs.

AI assistance supported archive inspection, analysis-code drafting, mathematical checking, visualization and language revision. Numerical claims remain bound to saved inputs, outputs and fixed certificates. Visible model answers are treated as behavioral observations.

\subsection*{Figure assembly and source data}
Statistical plots are regenerated from the saved paired records. Multipanel figures combine related measurements with lowercase sequential panel labels. Native screenshots remain unchanged embedded image layers. Coordinates, overlays, axes and labels are separate editable vector objects. Equal world-unit scales are used for geometric scene panels. The candidate-difference panel uses its explicitly labelled magnified vertical scale. Native hit displays stop at first sampled target entry, and the full stored CSV is retained. Figure-specific CSV and JSON files contain the underlying identities, exact values and provenance. No smoothing of the dense native paths or inferred model answer is introduced.

Additional experimental-process reconstructions are supplied in Supplementary Figs. 4-10. The companion atlas retains the final actual turn of all 30 completed episodes, and the offline replay animates stored trajectory samples. Every displayed request, answer and path is bound to archived source records. Separate matched requests are identified explicitly when arranged for comparison.

\section*{Data availability}
Source data underlying all six main figures accompany this author-review manuscript in source\_data\_nmi. The reproducibility package additionally contains 2,333 response-level records, frozen-world data, exact endpoint diagnostics, 72 catalog certificates and the 435 native turn events. Original selected screenshots and dense trajectory CSVs are supplied with SHA-256 bindings. The earlier 72-game pilot and independent-control raw archives remain outstanding. A permanent public accession for the new scientific analysis package is awaiting author deposit.
\section*{Code availability}
The analysis, figure-generation and validation scripts accompany the reproducibility package. The framework development release is available in the GAGR-Lab repository under tag v4.13.1-dev.3 \cite{ref8}. This analysis additionally binds its source data to the saved VGAGR4131\_20261105\_1 scoring epoch. Full raw-archive reconstruction uses the source framework and historical archives identified in the package. The supplied curated-data checks and figure-source tables can be inspected independently.
\section*{Acknowledgements and funding}
Funding and acknowledgement information is awaiting author confirmation.
\section*{Author contributions}
Individual author contributions are awaiting confirmation by the three authors.
\section*{Competing interests}
The competing interests declaration is awaiting author confirmation.

\clearpage
\pdfbookmark[0]{Experiment Process Atlas}{part.process}
\section*{Experiment Process Atlas}
\begin{center}
\includegraphics[page=1,width=\linewidth,height=0.88\textheight,keepaspectratio]{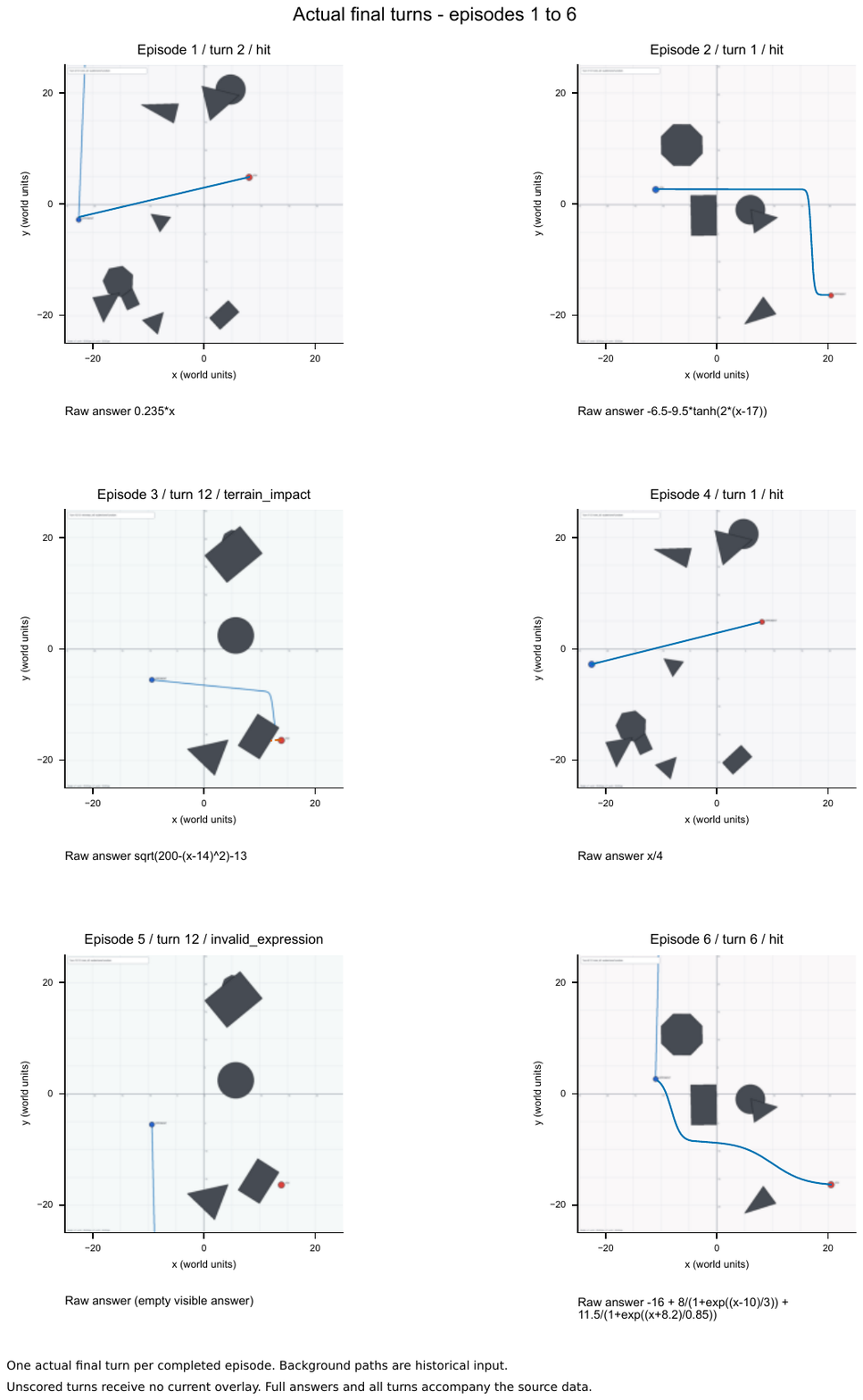}
\end{center}
\clearpage
\begin{center}
\includegraphics[page=2,width=\linewidth,height=0.97\textheight,keepaspectratio]{figures/Experiment_Process_Atlas.pdf}
\end{center}
\clearpage
\begin{center}
\includegraphics[page=3,width=\linewidth,height=0.97\textheight,keepaspectratio]{figures/Experiment_Process_Atlas.pdf}
\end{center}
\clearpage
\begin{center}
\includegraphics[page=4,width=\linewidth,height=0.97\textheight,keepaspectratio]{figures/Experiment_Process_Atlas.pdf}
\end{center}
\clearpage
\begin{center}
\includegraphics[page=5,width=\linewidth,height=0.97\textheight,keepaspectratio]{figures/Experiment_Process_Atlas.pdf}
\end{center}

\clearpage
\pdfbookmark[0]{Supplementary Information}{part.supplementary}
\section*{Supplementary Information}
\noindent Identifiable measurements of geometric action reasoning\par
\noindent Jingyao Zhang, Yun Li, Lu Han\par
\setcounter{figure}{0}
\setcounter{table}{0}
\renewcommand{\theHfigure}{supp.\arabic{figure}}
\renewcommand{\theHtable}{supp.\arabic{table}}
\renewcommand{\thefigure}{S\arabic{figure}}
\renewcommand{\thetable}{S\arabic{table}}
\section*{Supplementary methods and source accounting}
Supplementary Tables 1-3 give archive scoring states, full observed profiles and native cohort counts. Supplementary Figs. 1-3 and Tables 4-6 retain all seven turns in the three selected episodes. Supplementary Table 7 lists prospective interventions. Supplementary Table 8 separates earlier reports with unlocated raw archives from the reconstructed datasets.

Supplementary Figs. 4-10 reconstruct additional experimental processes from archived records. They show sample-index trajectory prefixes, all 12 actions of a completed zero-hit episode, actual matched formalization and synthesis answers, paired support prompts and outputs, four saved scoring-stage outcomes, an exact obstacle counterexample, and a complete image-plus-symbolic request. These reconstructions expose inputs, outputs and execution evidence. They do not observe internal model reasoning. The companion Experiment Process Atlas displays the final actual turn of every one of the 30 completed episodes. The offline replay animates the seven originally selected turns using stored sample indices and an editorial frame duration.

Qualification contributes 60 main-set requests, including 45 interface-qualification and 15 syntax qualification requests. Exploration executes 751/1,988 planned jobs. The later frozen stage executes 530/2,016. Their 14-config complete judgment panels contain five and one shared scenes, respectively. Missing planned calls remain missing. No profile is removed because its observed scores reach the floor or ceiling.

The following three cases include all seven turns, retaining the exact visible answer and historical executed expression. All seven turns have valid archived trajectories. Four miss the target and three record hits. These cases are selected within hit episodes. The complete cohort denominator is the separate 30-game, 435-action archive. Historical screenshots can already contain the previous trajectory, which remains part of the original visible input.

Episode \nolinkurl{20260902T161443Z_tournament_minimax_m3_vs_kimi_k3_A_left_A_first_ec390a4a4ae9a77e_a34e81ca}. Seed 20260912. The episode has 2 actions and 1 recorded hit.

CSV point counts in turn order are 1253, 3268. The corresponding logged minimum\_\allowbreak target\_\allowbreak distance values are 29.76431945248934, -0.03415748144128672. Full per-turn screenshot and CSV SHA-256 hashes, raw event fields and coordinate arrays are retained in native\_cases/native\_01/Case\_Evidence.json.

Episode \nolinkurl{20260910T155540Z_tournament_muse_glimmer_30b_vs_gemma_4_31b_A_right_B_first_476c9281f2c4c367_e8a70783}. Seed 20260922. The episode has 4 actions and 1 recorded hit.

CSV point counts in turn order are 4809, 4198, 896, 4189. The corresponding logged minimum\_\allowbreak target\_\allowbreak distance values are 0.5584515166704285, 0.33002552705712623, 38.975408969540545, -0.2506859465705385. Full per-turn screenshot and CSV SHA-256 hashes, raw event fields and coordinate arrays are retained in native\_cases/native\_02/Case\_Evidence.json.

Episode \nolinkurl{20260910T105115Z_tournament_muse_glimmer_30b_vs_kimi_k3_A_right_B_first_6a895eebdf73453e_6fbaaaa7}. Seed 20260914. The episode has 1 actions and 1 recorded hit.

CSV point counts in turn order are 4711. The corresponding logged minimum\_\allowbreak target\_\allowbreak distance values are -0.4473561983167152. Full per-turn screenshot and CSV SHA-256 hashes, raw event fields and coordinate arrays are retained in native\_cases/native\_03/Case\_Evidence.json.

The present data support operational descriptions of behavior. A model-internal universal deficit would require additional evidence about invariance across input representations, budgets and scene structures, together with interventions on the proposed reasoning stages. Related work on self-verification supplies important comparison conditions \cite{suppref5,suppref6}.

The original ZIP contains the first-draft TeX, bibliography files and a trajectory PDF. Its SHA-256 is \nolinkurl{8dc92becfd9ccd4809ad1d5b760e6ec894d459b3e5559ff4112527dc1ae4d4ec}. Those member bytes are preserved in source\_manuscript. The current quadratic certificate applies to continuous quadratic graphs with the stated coefficient and span. A prefix such as x squared in a reported answer is insufficient on its own to certify the complete function family. Locating and parsing every original response is required before applying the bound to the full pilot.

The response-level tables bind every reconstructed attempt to its response and score-record hashes. The delivery manifest additionally binds the plots, selected screenshot and trajectory bytes, analysis scripts, and manuscript source. All analyses and supplementary controls in this revision use saved observations or offline mathematics. They add zero model API calls.

\clearpage
\begin{figure}[htbp]\centering
\includegraphics[width=180mm]{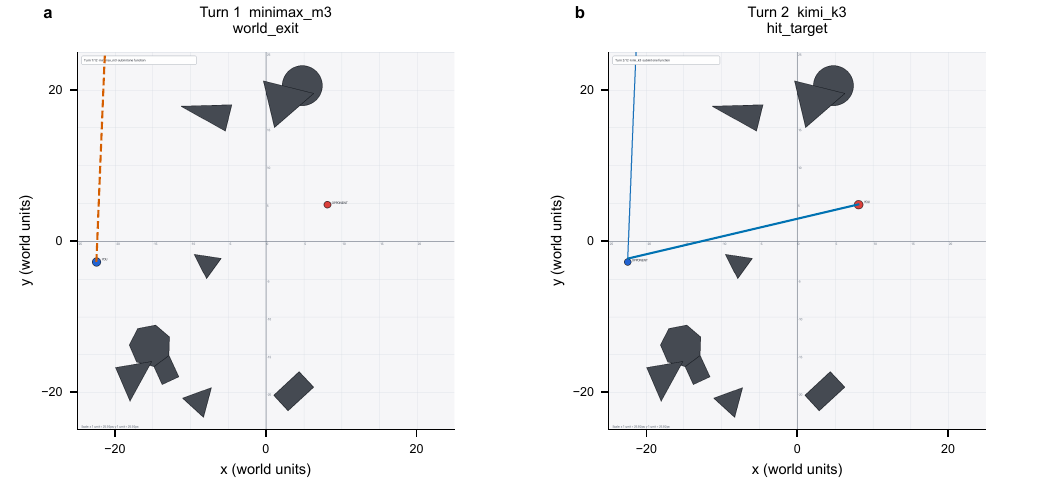}
\caption{Complete selected native episode 1. Panels a,b show all 2 archived turns in chronological order. Every panel uses its actual input screenshot. Dashed orange denotes a failed current action and solid blue denotes the current hit prefix. Previously visible trajectories remain in the original screenshot layer. The exact model answers and executed expressions are given in Supplementary Table 4. Full CSV paths and screenshot hashes accompany the source data.}
\end{figure}
\clearpage
\begin{figure}[htbp]\centering
\includegraphics[width=180mm]{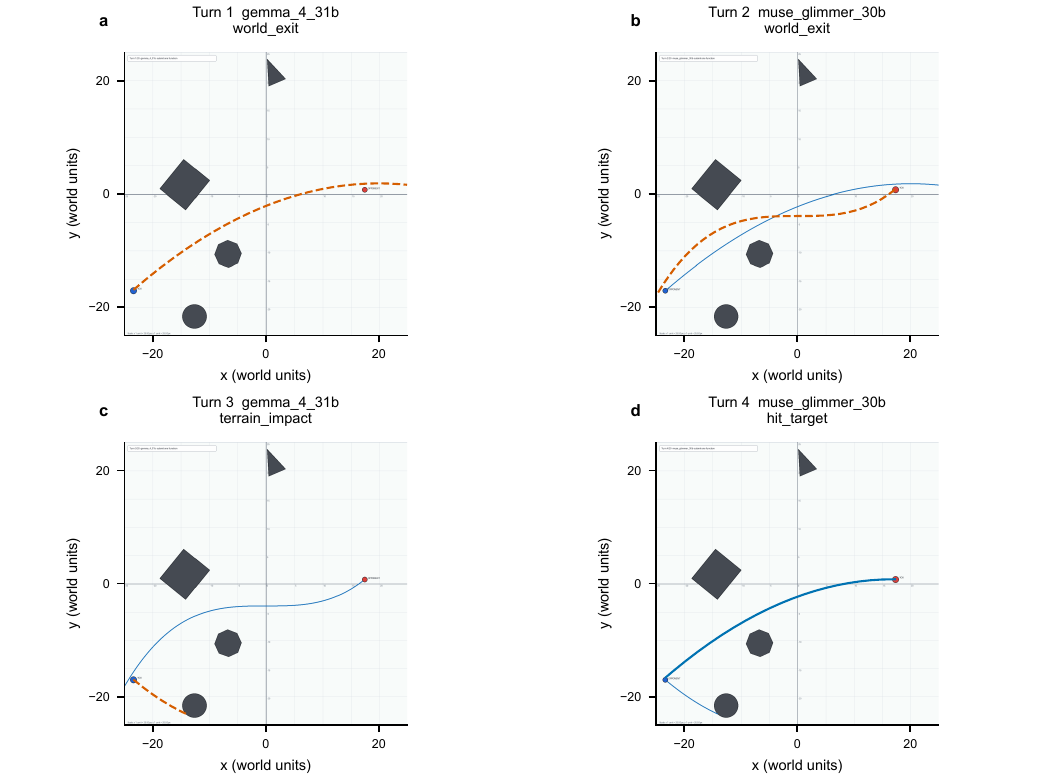}
\caption{Complete selected native episode 2. Panels a-d show all 4 archived turns in chronological order. Every panel uses its actual input screenshot. Dashed orange denotes a failed current action and solid blue denotes the current hit prefix. Previously visible trajectories remain in the original screenshot layer. The exact model answers and executed expressions are given in Supplementary Table 5. Full CSV paths and screenshot hashes accompany the source data.}
\end{figure}
\clearpage
\begin{figure}[htbp]\centering
\includegraphics[width=89mm]{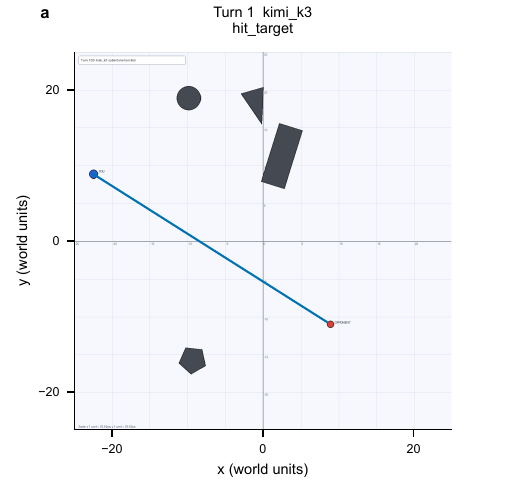}
\caption{Complete selected native episode 3. Panels a show all 1 archived turns in chronological order. Every panel uses its actual input screenshot. Dashed orange denotes a failed current action and solid blue denotes the current hit prefix. Previously visible trajectories remain in the original screenshot layer. The exact model answers and executed expressions are given in Supplementary Table 6. Full CSV paths and screenshot hashes accompany the source data.}
\end{figure}
\clearpage
\begin{figure}[htbp]\centering
\includegraphics[width=180mm]{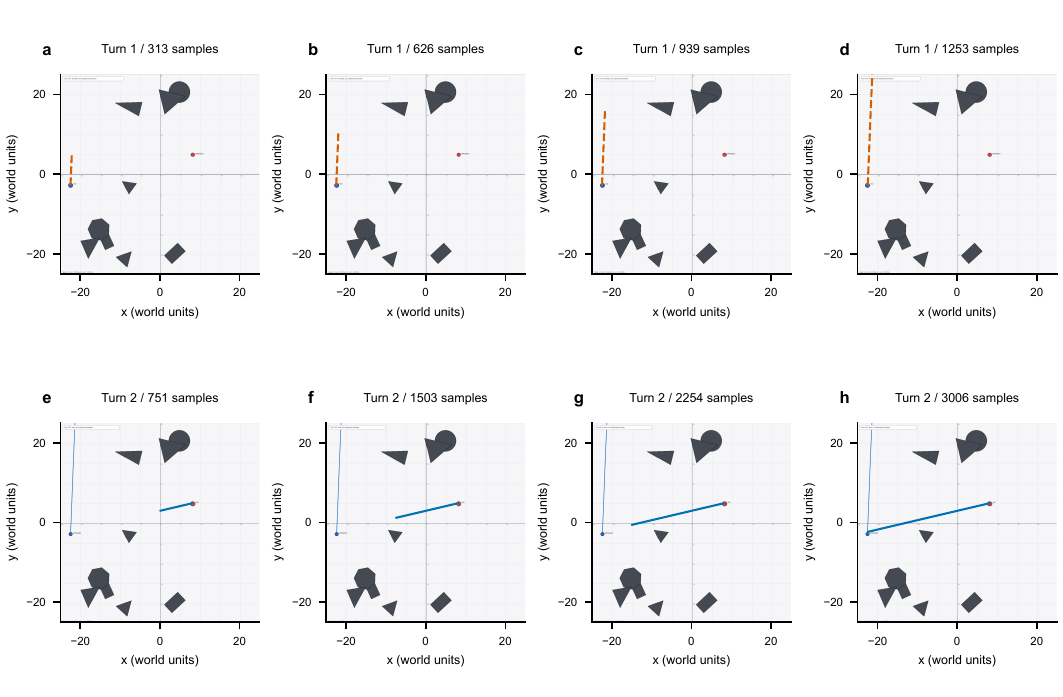}
\caption{Stored trajectory prefixes in a complete two-turn native episode. Panels a-d show the first action and e-h show the second at 25\%, 50\%, 75\% and 100\% of the displayed sample prefix. Every panel uses that turn's actual input screenshot. Dashed orange marks the failed sine action and solid blue the recorded linear hit. Sample index defines this reconstruction. The panels do not estimate physical time or reveal internal model reasoning.}
\end{figure}
\clearpage
\begin{figure}[htbp]\centering
\includegraphics[width=180mm]{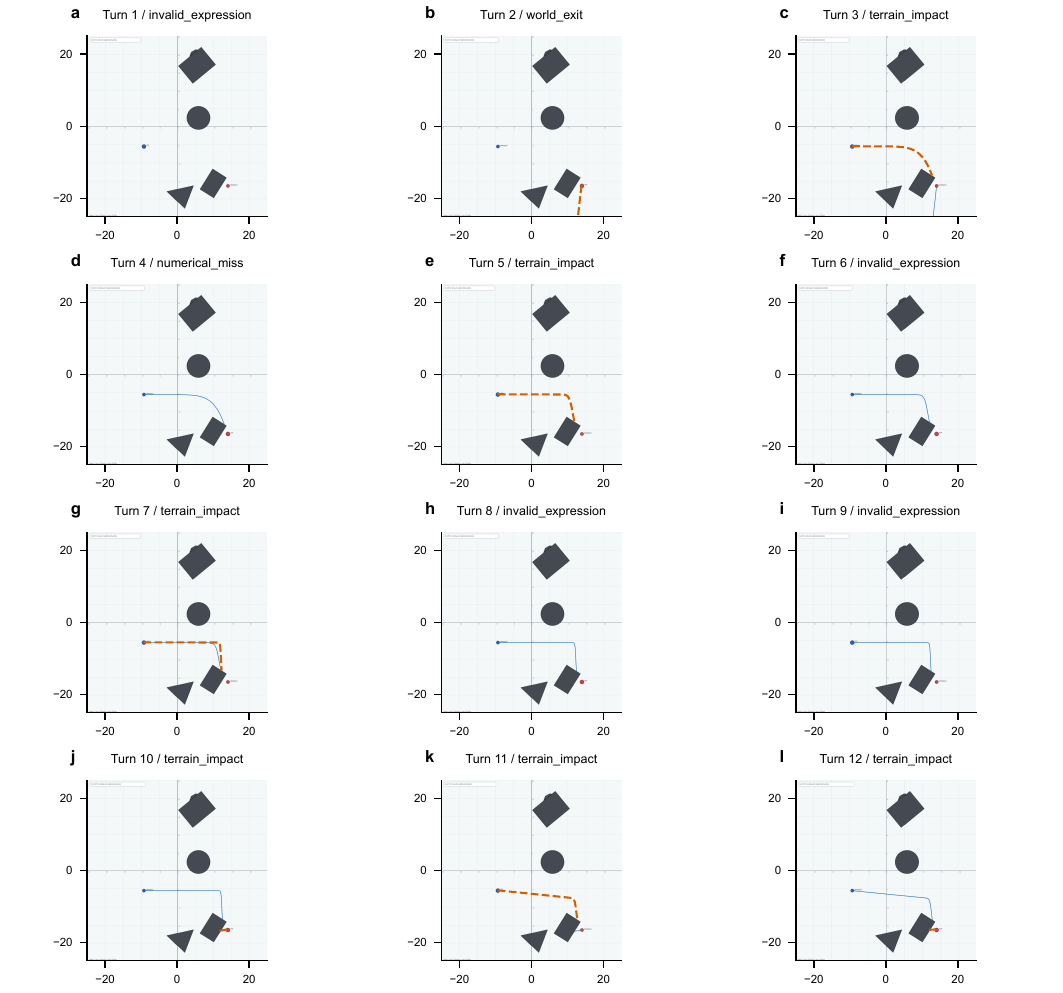}
\caption{All 12 attempts in the earliest completed zero-hit episode in the first historical cohort. Panels a-l follow archived turn order, using each actual screenshot and its stored current-action path when one exists. The scene retains a common equal-scale world-coordinate frame. Background trajectories are part of the original observations. The source data include every raw answer, executed expression and outcome for this failed episode.}
\end{figure}
\clearpage
\begin{figure}[htbp]\centering
\includegraphics[width=180mm]{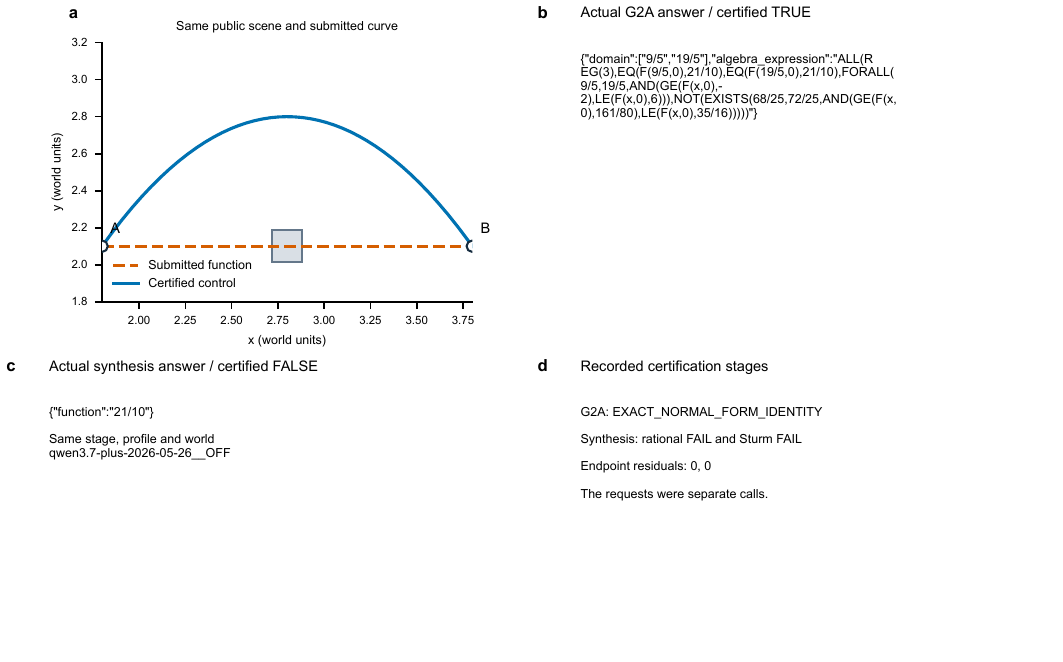}
\caption{An actual matched formalization-success and synthesis-failure pair. a, The public geometry, submitted constant function and certified feasible control. b,c, The archived model answers. d, Saved fixed-algorithm certification fields. Both requests share stage, profile and world, and were sent separately. The displayed analytical order is distinct from a model-generated chained workflow. Full request text, response identities, hashes and certificates are supplied.}
\end{figure}
\clearpage
\begin{figure}[htbp]\centering
\includegraphics[width=180mm]{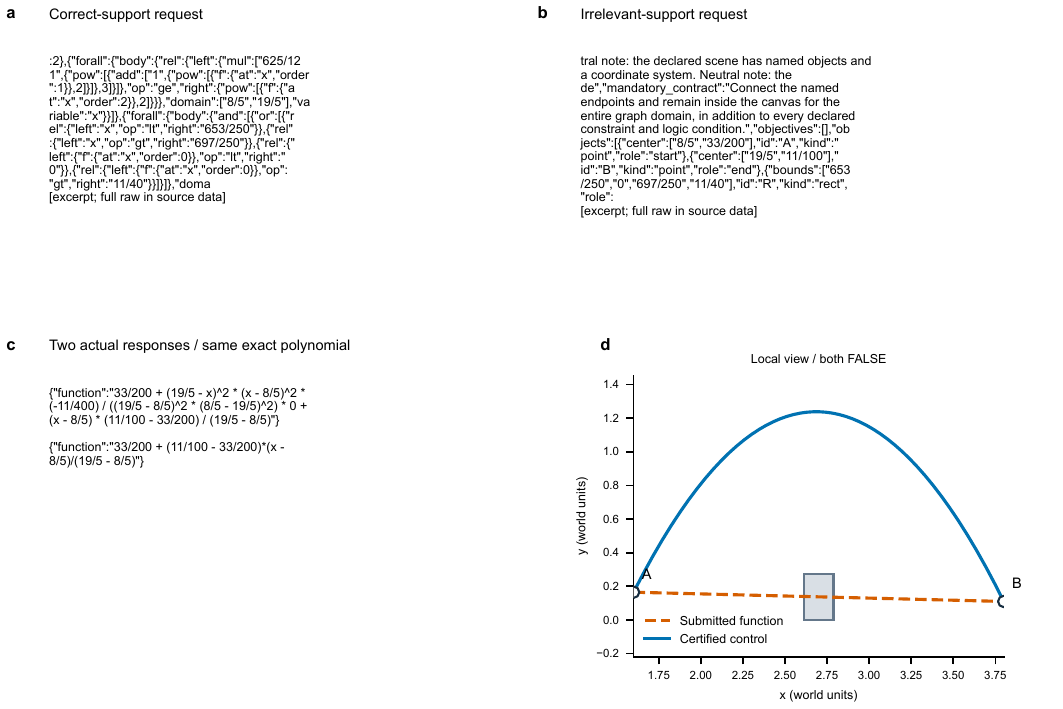}
\caption{Actual correct-support and irrelevant-support requests in one matched pair. a,b, Line-wrapped excerpts from the saved prompt tails, with full text in the source record. c, Both complete visible responses. d, Their common exact polynomial graph and a certified feasible control. The two response strings differ while their polynomial coefficients agree exactly. Both saved synthesis judgments are FALSE. This case describes observable outputs and does not establish whether the model internally used the support.}
\end{figure}
\clearpage
\begin{figure}[htbp]\centering
\includegraphics[width=180mm]{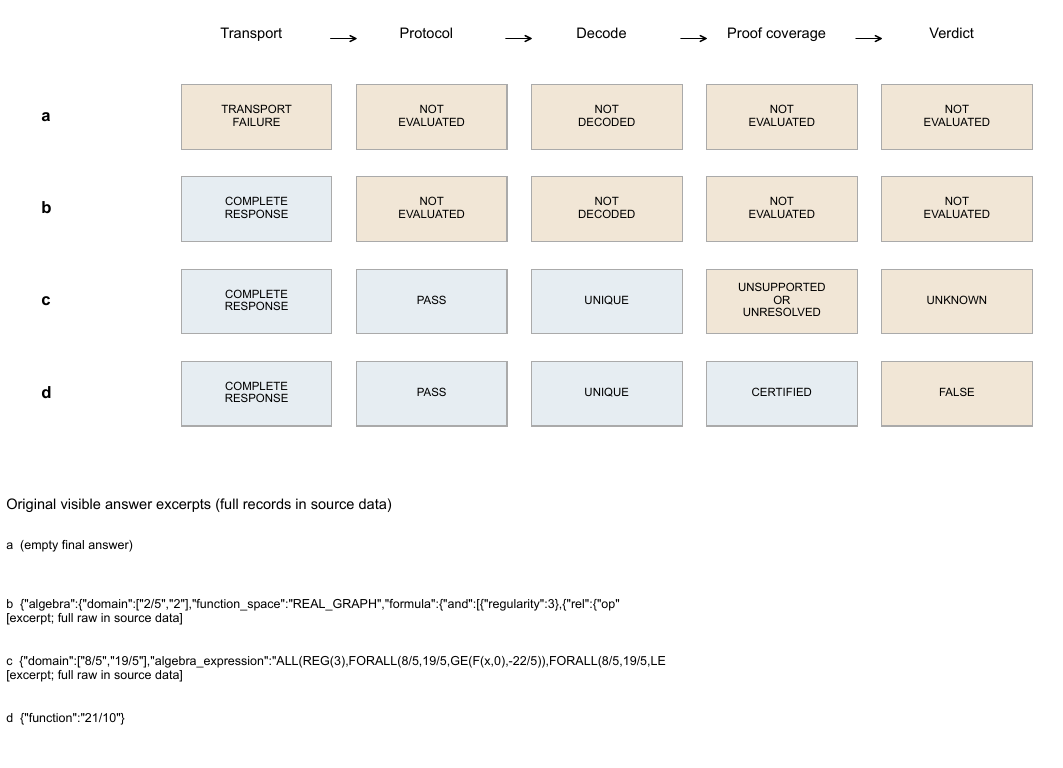}
\caption{Four actual responses followed through the saved scoring stages. a, A transport failure with no final visible answer. b, An incomplete or invalid JSON response. c, A decoded formalization without a certified equivalence result. d, A certified synthesis failure. Fields are copied from the fixed scoring record. The panels retain NOT\_EVALUATED, UNKNOWN and FALSE as distinct outcomes. Full original visible answers and unabridged score objects accompany the source data.}
\end{figure}
\clearpage
\begin{figure}[htbp]\centering
\includegraphics[width=180mm]{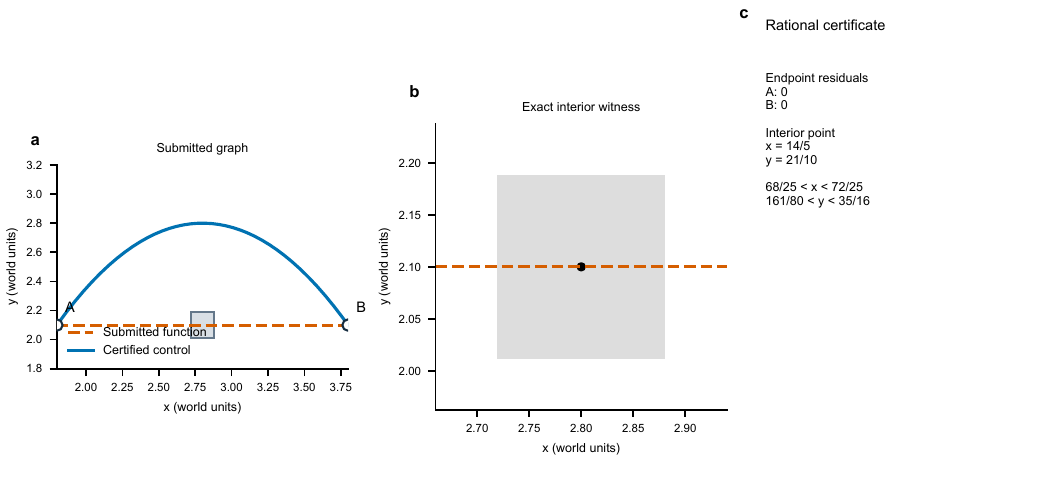}
\caption{A local exact explanation of one whole-domain synthesis failure. a, The submitted constant graph satisfies both endpoints. b, The rational point at the rectangle midpoint lies strictly inside the obstacle and on the submitted graph. c, Exact endpoint residuals and interval inequalities. This fixed arithmetic counterexample is reconstructed from the saved expression and public geometry. It supplies a checkable mathematical reason for the archived FALSE result.}
\end{figure}
\clearpage
\begin{figure}[htbp]\centering
\includegraphics[width=180mm]{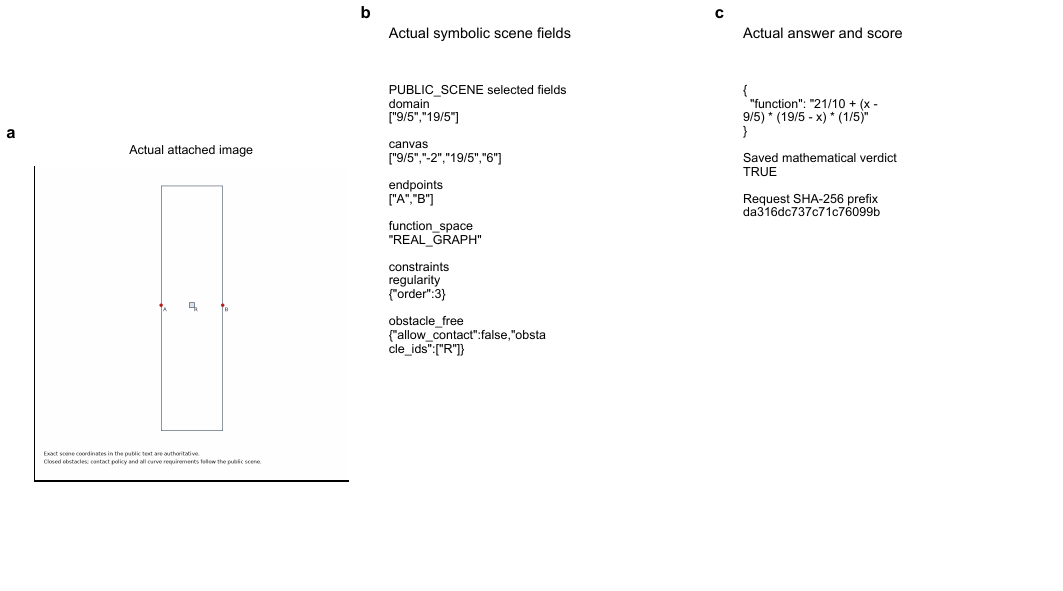}
\caption{An actual image-arm request and its response. a, The exact PNG attachment decoded from the frozen request body. b, Selected fields parsed from its complete symbolic PUBLIC\_SCENE text and laid out for readability. c, The original visible model answer and saved mathematical verdict. The request-body hash is matched to the archived response. This observation contains both visual and exact symbolic information. The full prompt, image bytes, answer and certificate are supplied in the source data.}
\end{figure}
\clearpage
\section*{Supplementary tables}
\begin{table}[htbp]\centering\small\setlength{\tabcolsep}{4pt}
\caption{Archived scoring states under the saved regrading epoch. TRUE/FALSE refer to correctness under each operation contract. Historical native actions are analyzed separately.}
\begin{tabular}{>{\raggedright\arraybackslash}p{0.267\linewidth}>{\raggedright\arraybackslash}p{0.089\linewidth}>{\raggedright\arraybackslash}p{0.098\linewidth}>{\raggedright\arraybackslash}p{0.098\linewidth}>{\raggedright\arraybackslash}p{0.187\linewidth}>{\raggedright\arraybackslash}p{0.151\linewidth}}
\toprule Call archive & N & TRUE & FALSE & Not evaluated & Unknown \\
Main calls & 1341 & 673 & 570 & 77 & 21 \\
Historical calls & 968 & 558 & 347 & 63 & 0 \\
Complex G2A calls & 24 & 5 & 8 & 10 & 1 \\
\bottomrule\end{tabular}\end{table}
\begin{table}[htbp]\centering\small\setlength{\tabcolsep}{4pt}
\caption{Observed exploration and later-stage serving profiles. Requested labels retain provider metadata scope. Every row has a TEXT\_ONLY response-carrier setting. The image-arm input condition is audited separately.}
\begin{tabular}{>{\raggedright\arraybackslash}p{0.525\linewidth}>{\raggedright\arraybackslash}p{0.134\linewidth}>{\raggedright\arraybackslash}p{0.098\linewidth}>{\raggedright\arraybackslash}p{0.134\linewidth}}
\toprule Requested model label & Mode & Temp. & Token cap \\
deepseek-v4.1-flash-1017 & DEFAULT & 0 & 16384 \\
deepseek-v4.1-flash & DEFAULT & 0 & 16384 \\
kimi-k2.6 & OFF & 0 & 16384 \\
kimi-k2.6 & ON & 0 & 16384 \\
kimi-k2.7-code & DEFAULT & 0 & 16384 \\
kimi-k3 & DEFAULT & 1.0 & 16384 \\
qwen3.5-flash-2026-02-23 & OFF & 0 & 16384 \\
qwen3.5-flash-2026-02-23 & ON & 0 & 16384 \\
qwen3.7-flash-2026-07-15 & OFF & 0 & 16384 \\
qwen3.7-flash-2026-07-15 & ON & 0 & 16384 \\
qwen3.7-max-2026-06-08 & OFF & 0 & 16384 \\
qwen3.7-max-2026-06-08 & ON & 0 & 16384 \\
qwen3.7-plus-2026-05-26 & OFF & 0 & 16384 \\
qwen3.7-plus-2026-05-26 & ON & 0 & 16384 \\
\bottomrule\end{tabular}\end{table}
\begin{table}[htbp]\centering\small\setlength{\tabcolsep}{4pt}
\caption{Historical cohort counts reconciled against raw turn events. H1 requests Minimax/Kimi, H2 requests Muse/Gemma under the syntax-v2 policy, and H3 requests Muse/Kimi. These cohorts have different protocols.}
\begin{tabular}{>{\raggedright\arraybackslash}p{0.285\linewidth}>{\raggedright\arraybackslash}p{0.134\linewidth}>{\raggedright\arraybackslash}p{0.151\linewidth}>{\raggedright\arraybackslash}p{0.178\linewidth}>{\raggedright\arraybackslash}p{0.142\linewidth}}
\toprule Native cohort & Games & Actions & Valid paths & Hits \\
H1 & 6 & 34 & 26 & 4 \\
H2 & 12 & 180 & 141 & 4 \\
H3 & 12 & 221 & 9 & 1 \\
\bottomrule\end{tabular}\end{table}
\begin{table}[htbp]\centering\small\setlength{\tabcolsep}{4pt}
\caption{Exact raw answers and historical extraction results for native\_01.}
\begin{tabular}{>{\raggedright\arraybackslash}p{0.214\linewidth}>{\raggedright\arraybackslash}p{0.303\linewidth}>{\raggedright\arraybackslash}p{0.374\linewidth}}
\toprule Turn / actor & Raw visible answer & Executed expression / outcome \\
1 / minimax\_m3 & y = -25.92*sin(x) & -25.92*sin(x) / world\_exit \\
2 / kimi\_k3 & 0.235*x & 0.235*x / hit\_target \\
\bottomrule\end{tabular}\end{table}
\begin{table}[htbp]\centering\small\setlength{\tabcolsep}{4pt}
\caption{Exact raw answers and historical extraction results for native\_02.}
\begin{tabular}{>{\raggedright\arraybackslash}p{0.214\linewidth}>{\raggedright\arraybackslash}p{0.303\linewidth}>{\raggedright\arraybackslash}p{0.374\linewidth}}
\toprule Turn / actor & Raw visible answer & Executed expression / outcome \\
1 / gemma\_4\_31b & -0.01*x\textasciicircum{}2 + 0.4*x + 20 & -0.01*x\textasciicircum{}2 + 0.4*x + 20 / world\_exit \\
2 / muse\_glimmer\_30b & 0.000906*x\textasciicircum{}3 & 0.000906*x\textasciicircum{}3 / world\_exit \\
3 / gemma\_4\_31b & 0.02*x\textasciicircum{}2+0.1*x & 0.02*x\textasciicircum{}2+0.1*x / terrain\_impact \\
4 / muse\_glimmer\_30b & 1-18*(x-17)\textasciicircum{}2/1681 & 1-18*(x-17)\textasciicircum{}2/1681 / hit\_target \\
\bottomrule\end{tabular}\end{table}
\begin{table}[htbp]\centering\small\setlength{\tabcolsep}{4pt}
\caption{Exact raw answers and historical extraction results for native\_03.}
\begin{tabular}{>{\raggedright\arraybackslash}p{0.214\linewidth}>{\raggedright\arraybackslash}p{0.303\linewidth}>{\raggedright\arraybackslash}p{0.374\linewidth}}
\toprule Turn / actor & Raw visible answer & Executed expression / outcome \\
1 / kimi\_k3 & -0.633*x-5.31 & -0.633*x-5.31 / hit\_target \\
\bottomrule\end{tabular}\end{table}
\begin{table}[htbp]\centering\small\setlength{\tabcolsep}{4pt}
\caption{Prospective hypotheses and tests. These interventions have zero newly executed model observations in the present study.}
\begin{tabular}{>{\raggedright\arraybackslash}p{0.205\linewidth}>{\raggedright\arraybackslash}p{0.356\linewidth}>{\raggedright\arraybackslash}p{0.329\linewidth}}
\toprule Question & Controlled intervention & Discriminating observation \\
Transfer from relations to action & Replace the generated formal object with a certified one within the same chained scene. & A persistent construction deficit would localize failure downstream of that intermediate object. \\
Local and global judgment & Keep endpoints and all non-target properties fixed while violating exactly one global requirement. & Recheck whether judgment patterns persist when endpoint-only rejection is unavailable. \\
Use of symbolic support & Control token length, output budget and redundant scene information across support arms. & Separate useful representation changes from redundant or unused information. \\
Necessity of visual information & Hold text fixed and change images so admissible answer sets differ. & Measure sensitivity to information that the scored operation actually requires. \\
Difficulty and generalization & Freeze base-family splits and introduce certified single-factor ladders with low/high anchors. & Test whether capability ordering and operation separation replicate on held-out structures. \\
\bottomrule\end{tabular}\end{table}
\begin{table}[htbp]\centering\small\setlength{\tabcolsep}{4pt}
\caption{Numbers retained as earlier manuscript reports. They are excluded from the newly reconstructed 2,333-call and 30-game denominators.}
\begin{tabular}{>{\raggedright\arraybackslash}p{0.249\linewidth}>{\raggedright\arraybackslash}p{0.356\linewidth}>{\raggedright\arraybackslash}p{0.285\linewidth}}
\toprule Earlier first-draft report & Reported outcome & Current source status \\
Llama pilot & 72 games, 432 attempts, 429 valid ordinary responses, zero hits. & Original raw directory remains to be located. \\
One-shot prompt diagnostics & 72 valid ordinary-function responses, zero hits. & Original raw directory remains to be located. \\
Independent analytic control & 600 directional cases succeed, with 600 repeat and 1,200 transform checks. & Raw independent-control archive remains to be located. \\
\bottomrule\end{tabular}\end{table}

\end{document}